%% file: main.tex
\RequirePackage{amsmath,amssymb,amsfonts}
\documentclass[runningheads]{llncs}
\usepackage[T1]{fontenc}
\usepackage{graphicx}
\usepackage{orcidlink}
\renewcommand{\orcidID}[1]{\,\orcidlink{#1}}
\usepackage{cite}
\usepackage{amsmath,amssymb,amsfonts}
\usepackage{algorithmic}
\usepackage{textcomp}
\usepackage{xcolor}
\usepackage[capitalise]{cleveref}
\usepackage{caption}
\usepackage{subcaption}
\usepackage{url}
\usepackage{booktabs}
\usepackage{tabularx}
\usepackage{multirow}
\usepackage{multicol}
\usepackage{pseudo}
\usepackage{hyperref}
\usepackage{tikz}
\pseudodefinestyle{fullwidth}{
  begin-tabular = \tabularx{\linewidth}{@{}
    r 
    >{\pseudosetup} 
    X 
    >{\leavevmode\small\color{black!60}} 
    p{,\linewidth} 
    @{}},
  end-tabular = \endtabularx,
  setup-append = \pseudoeq
}
\pseudoset{
  ctfont=\color{black!55}\textit,
  ct-left=//\ ,
  ct-right=,
}
\newcommand{\matr}[1]{\mathbf{#1}}
\usepackage{float}
\usepackage{comment}
\usepackage{makecell}
\usepackage{enumitem}

\begin{document}
\title{Accelerating Mobile Inference through Fine-Grained CPU-GPU Co-Execution}
\titlerunning{Accelerating Mobile Inference through CPU-GPU Co-Execution}
%
\author{
Zhuojin Li\orcidID{0000-0002-8308-0231} \and
Marco Paolieri\orcidID{0000-0001-5110-203X} \and
Leana Golubchik\orcidID{0000-0001-8353-5040}
}
\authorrunning{Li et al.}
%
\institute{University of Southern California, Los Angeles, California, USA \\
\email{\{zhuojinl,paolieri,leana\}@usc.edu}}
\maketitle              
\begin{tikzpicture}[remember picture,overlay]
\node[anchor=north west,shift={(0,0)},minimum width=\paperwidth,minimum height=3.3cm,fill=black!5] at (current page.north west){
\begin{minipage}[t]{0.15\paperwidth}
\vspace{-2mm}%
\raggedleft BibTeX Citation:
\end{minipage}\hspace{5mm}
\begin{minipage}[t]{0.8\paperwidth}\footnotesize
\vspace{-2mm}%
\begin{verbatim}
@inproceedings{LiPG26b,
  author    = {Zhuojin Li and Marco Paolieri and Leana Golubchik},
  title     = {Accelerating Mobile Inference through Fine-Grained {CPU}-{GPU} Co-Execution},
  booktitle = {Selected Papers of {EPEW} 2025},
  series    = {Lecture Notes in Computer Science}, volume = {15657}, pages = {41--55},
  publisher = {Springer}, year = {2026}, doi = {10.1007/978-3-032-16345-5\_4}
}
\end{verbatim}
\end{minipage}
};
\end{tikzpicture}
\begin{abstract}
\input{sections/0-abstract}

\end{abstract}

\input{sections/1-intro}

\input{sections/2-partition}

\input{sections/3-predictor}

\input{sections/4-overhead}

\input{sections/5-results}

\input{sections/6-conclusion}
%
%
%
\bibliographystyle{splncs04}
\bibliography{main}

\end{document}

%% file: sections/0-abstract.tex
Deploying deep neural networks on mobile devices is increasingly important but remains challenging due to limited computing resources.
On the other hand, their unified memory architecture and narrower gap between CPU and GPU performance provide an opportunity to reduce inference latency by assigning tasks to both CPU and GPU.
The main obstacles for such collaborative execution are the significant synchronization overhead required to combine partial results, and the difficulty of predicting execution times of tasks assigned to CPU and GPU (due to the dynamic selection of implementations and parallelism level).
To overcome these obstacles, we propose both a lightweight synchronization mechanism based on OpenCL fine-grained shared virtual memory (SVM) and machine learning models to accurately predict execution times.
Notably, these models capture the performance characteristics of GPU kernels and account for their dispatch times.
A comprehensive evaluation on four mobile platforms shows that our approach can quickly select CPU-GPU co-execution strategies achieving up to 1.89x speedup for linear layers and 1.75x speedup for convolutional layers (close to the achievable maximum values of 2.01x and 1.87x, respectively, found by exhaustive grid search on a Pixel~5 smartphone).

%% file: sections/1-intro.tex
\section{Introduction}\label{section:introduction}

Machine learning (ML) techniques have achieved rapid growth in recent years, driven by the breakthroughs in large-scale architectures such as Large Language Models (LLMs) and Large Vision Models (LVMs). These state-of-the-art models exhibit impressive capabilities in a wide range of applications, including question answering, image recognition, and video analysis.
While these models are typically trained on powerful cloud servers, there is a growing demand for deployment on mobile platforms to serve inference tasks, since on-device deployment not only protects user privacy, but also provides offline availability and reduces latency for real-time applications.

However, deploying large-scale models on mobile platforms remains challenging, due to limited computational resources and energy constraints.
%
To improve inference latency on mobile platforms, substantial research and industrial efforts have focused on developing efficient neural network architectures~\cite{sandler2018mobilenetv2} and specialized hardware~\cite{jang2021sparsity}.
On top of these efforts, our work explores an additional dimension: distributing fine-grained operations (e.g., forward propagation of a neural network layer) across heterogeneous computing devices (specifically, CPU and GPU) available on modern mobile platforms.
In particular, we pursue this approach by leveraging the following opportunities.

\begin{figure}[t]
    \centering
    \begin{subfigure}[t]{.635\linewidth}
        \centering
        \includegraphics[width=\linewidth]{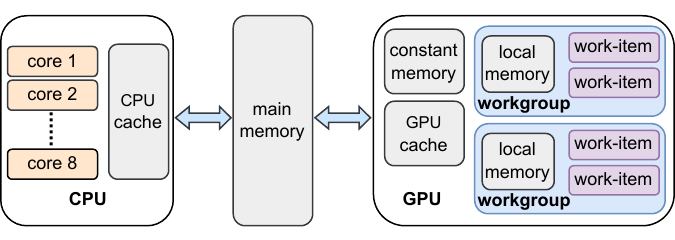}
        \caption{Unified Memory}\label{fig:intro_mobile_soc}
    \end{subfigure}
    \hspace{-.5em}
    \begin{subfigure}[t]{.353\linewidth}
        \centering
        \includegraphics[width=\linewidth]{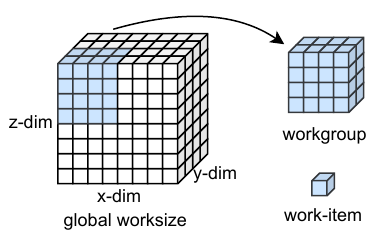}
        \caption{GPU Workgroup}\label{fig:intro_gpu_workgroup}
    \end{subfigure}
    \caption{Introduction to Mobile Platforms}\label{fig:introduction}
\end{figure}

The unified memory architecture on mobile platforms enables both the main processor (CPU) and specialized accelerators (e.g., GPU and NPU) to directly access shared areas of main memory, as illustrated in \cref{fig:intro_mobile_soc}.
In contrast, GPUs of cloud servers maintain separate on-chip memories and require data transfers to share data with the CPU through the main memory.
By eliminating the need for memory transfers to main memory, unified memory on mobile platforms facilitates collaborative execution of inference tasks across compute devices.


\begin{figure}[t]
    \centering
    \includegraphics[width=.75\linewidth]{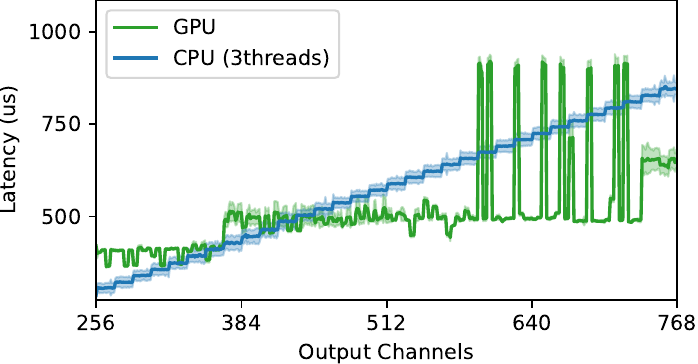}
    \caption{Comparison of CPU and GPU Latencies for Linear Operations with Input Shape (50, 3072), with 95\% Confidence Intervals (OnePlus~11)}\label{fig:motivation_CPU_GPU_gap}
\end{figure}

\begin{figure}[t]
        \centering
        \includegraphics[width=.75\linewidth]{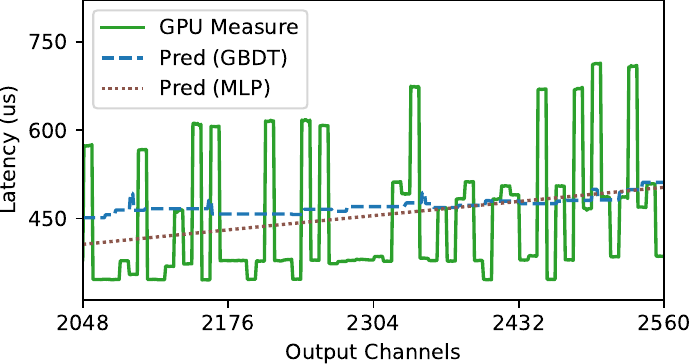}\vspace{-.5em}
        \caption{Inadequate Modeling of GPU Latency Spikes by Existing Methods for Linear Operations with Input Shape (50, 768) (OnePlus~11)}\label{fig:motivation_inaccurate_predictors}
\end{figure}

Moreover, in contrast with existing works~\cite{jayanth2024benchmarking,tang2021bridge} which report poor performance on mobile CPUs (due to inefficient CPU implementations in some ML libraries such as OpenVINO~\cite{openvino}), our empirical analysis shows that the performance gap between mobile CPUs and GPUs can be narrow for important operations when using the XNNPACK~\cite{xnnpack} library in TensorFlow Lite (TFLite), which provides high-performance implementations based on advanced SIMD instructions for ARM CPUs. For instance, as illustrated in \cref{fig:motivation_CPU_GPU_gap}, for matrix multiplications (i.e., a linear layer in a neural network) of sizes $50\times 3072$ and $3072\times C_{out}$, a CPU implementation with 3 threads achieves even lower latency than a GPU kernel when $C_{out} < 425$.
The significant performance of mobile CPUs motivates us to \emph{assign compute tasks to both CPU and GPU to achieve lower inference latency through parallel execution.}

However, designing an effective co-execution strategy across multiple accelerators remains challenging. In our empirical evaluation, existing approaches achieve limited performance gains mainly due to the following two reasons.

First, optimal workload partitioning across CPUs and GPUs is difficult, primarily because GPU kernels exhibit complex non-linear performance characteristics.
Our experiments, reported in \cref{fig:motivation_inaccurate_predictors}, show significant latency spikes (green line) for linear operations as the output dimension $C_{out}$ increases, due to heuristic choices in GPU kernel implementations and in the assignment of tasks to \emph{workgroups} (i.e., to groups of threads executing the same code on different subsets of the input data, as illustrated in \cref{fig:intro_gpu_workgroup}).
%
Consequently, co-execution frameworks relying on linear models for GPU latency prediction (e.g.,~\cite{chen2025heterollm}) can make poor partitioning decisions.
Although non-linear ML models based on operation configurations (e.g., input/output channels) have been proposed for more accurate latency predictions~\cite{jia2022codl,zhang2021nn,LiPG24b,LiPG23}, our evaluations in \cref{fig:motivation_inaccurate_predictors} indicate that these methods\footnote{GBDT hyperparameters are detailed in \cref{section:evalution_setup}. The MLP predictor was selected by varying the number of layers (1 to 4), neurons at each layer (32, 64 or 128), dropout rate (0 to 0.5), learning rate ($10^{-5}$ to $0.1$) and weight decay ($10^{-6}$ to $10^{-3}$).} still fail to capture sudden latency spikes for specific configurations.
%
%
%

Second, synchronization overhead between compute devices is often non-negligible.
Existing work~\cite{jia2022codl} reports overhead of up to 1~ms for the GPU to notify the CPU that data mapping of a shared memory region has completed; as a comparison, our measurements indicate that, using the GPU on the OnePlus~11 smartphone, the longest linear operation of the ViT-Base-32 neural network~\cite{dosovitskiy2020image} takes only 660~$\mu$s.
%
Such overhead can completely wipe out the benefits of co-execution; thus, minimizing synchronization overhead is crucial.

%
Our main contributions to tackling these problems are as follows:
\begin{itemize}
\item We develop accurate latency predictors that use kernel implementation details and kernel dispatch behaviors from TFLite (\cref{section:latency_prediction}).
Using detailed information, our predictors accurately capture complex latency characteristics, including the discontinuity due to heuristic workgroup choices and kernel selection, enabling more effective workload partitioning decisions.

\item To address significant synchronization overhead, we use fine-grained shared virtual memory in OpenCL to implement an efficient CPU-GPU synchronization mechanism on mobile platforms (\cref{section:sync_overhead}).
Our design reduces expensive data mapping operations required for cache coherence and avoids notification delay through active querying.
As a result, synchronization overhead is substantially reduced, e.g., from 162~$\mu$s to 7~$\mu$s for linear operations on a Motorola Edge Plus 2022 smartphone.
%


\item We create a comprehensive dataset consisting of latency measurements of 2,039 linear and 2,051 convolution operations executed on four mobile devices using co-execution strategies with 1 to 3 CPU threads and the GPU.
The evaluation shows that our predictors can quickly identify co-execution strategies achieving speedups up to 1.89x for linear operations and 1.75x for convolution operations (on Pixel~5 smartphones); we show that speedups are comparable to those identified through brute-force exploration of co-execution strategies (\cref{section:evalution}).

\end{itemize}

%% file: sections/2-partition.tex
\section{Inference Workload Partitioning}\label{section:background_partition}



\begin{figure}[t]
    \centering
    \begin{subfigure}[t]{.7\linewidth}
        \centering
        \includegraphics[width=\linewidth]{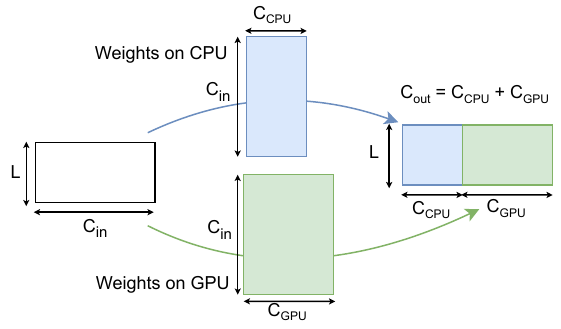}\vspace{-.5em}
        \caption{Linear Operation\vspace{1em}}\label{fig:partition_linear}
    \end{subfigure}
    \begin{subfigure}[t]{.7\linewidth}
        \centering
        \includegraphics[width=\linewidth]{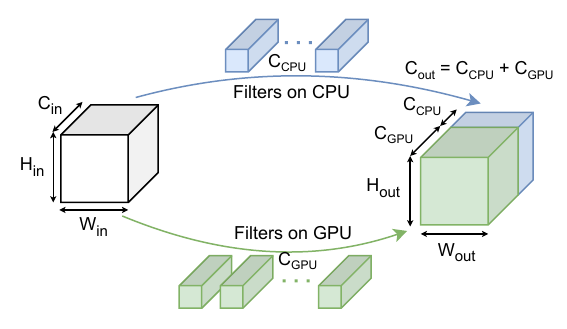}\vspace{-.5em}
        \caption{Convolution Operation}\label{fig:partition_conv}
    \end{subfigure}
    \caption{Illustration of Computation Partitioning over Output Channels}\label{fig:partition_details}
    \vspace{-1em}
\end{figure}

\paragraph{ML Operation Partitioning.}
Linear and convolutional layers are fundamental building blocks in deep neural networks.
A linear layer multiplies an input matrix $\matr{X} \in \mathbb{R}^{L\times C_{in}}$ (activations from the previous layer) by the \emph{weights} $\matr{W} \in \mathbb{R}^{C_{in}\times C_{out}}$ (the trainable parameters of the layer); each column of the output $\matr{Y}=\matr{X} \matr{W} \in  \mathbb{R}^{L\times C_{out}}$ represents a different output feature.
%
%
A convolutional layer is a generalization of a linear layer that is used to process data with grid-like topology (e.g., images): each output value $(\matr{Y})_{ijk}$ is obtained through the dot product of a \emph{kernel} (or \emph{filter}) $\matr{W}_k \in \mathbb{R}^{K \times K \times C_{in}}$ with a $K \times K$ patch of the input \emph{feature map} $\matr{X} \in \mathbb{R}^{H_{in} \times W_{in} \times C_{in}}$ (2D activations from a previous layer), i.e., $(\matr{Y})_{ijk} = \matr{X}(i,j)\matr{W}_k$ where $\matr{X}(i,j)$ restricts the first two dimensions of $\matr{X}$ to $[i-\lfloor\frac{K}{2}\rfloor, i+ \lfloor\frac{K}{2}\rfloor]$ and $[j-\lfloor\frac{K}{2}\rfloor, j+ \lfloor\frac{K}{2}\rfloor]$, respectively.
In the output $\matr{Y} \in \mathbb{R}^{H_{out} \times W_{out} \times C_{out}}$, input height and width can be reduced by using a \emph{stride} $S>1$ to increment $i$ and $j$, i.e., $H_{out} = \lfloor H_{in}/S \rfloor$ and $W_{out} = \lfloor W_{in}/S \rfloor$; the number of \textit{output channels} $C_{out}$ is equal to the number of different kernels $\matr{W}_k$.
%

In this work, we focus on partitioning the computation of linear or convolutional layers along output channels.
Since each output channel corresponds to a distinct column of a linear weight matrix~$\matr{W}$ or to a distinct convolution kernel~$\matr{W}_k$, each compute unit can store and manage its own subset of weights.
\cref{fig:partition_details} illustrates this strategy: the total number of output channels $C_{out}$ is partitioned as $C_{CPU} + C_{GPU} = C_{out}$. Accordingly, the original weights are split by assigning the first $C_{CPU}$ columns of $\matr{W}$ or the first $C_{CPU}$ kernels $\matr{W}_k$ to the CPU and the rest to the GPU. During execution, CPU and GPU calculate their assigned portion of output independently using the shared input $\matr{X}$.

%


\paragraph{Problem Formulation.}
%
Formally, given a total of $C_{out}$ output channels for a linear or convolution operation, our goal is to determine the optimal partitioning $c_1 + c_2 = C_{out}$ that minimizes the parallel execution time, i.e.,
$$
\min_{c_1 + c_2 = C_{out}} T_{overhead}(c_1,c_2) + \max\left(T_{CPU}(c_1), T_{GPU}(c_2)\right)\,.
$$
Here, $T_{CPU}(c_1)$ and $T_{GPU}(c_2)$ represent the computation latency on CPU and GPU, respectively. $T_{overhead}(c_1,c_2)$ accounts for additional latency from synchronization overhead; in particular, we observe that the overhead remains constant in our measurements when a co-execution strategy is used (\cref{section:sync_overhead}), while $T_{overhead}(c_1,c_2)=0$ when $c_1=C_{out}$ or $c_2=C_{out}$ (i.e., exclusive execution on CPU or GPU, respectively).
The formulation aims at balancing computational loads across CPU and GPU, ensuring high resource utilization.

Since the number of output channels can be thousands in practice (e.g., 3,072 in the vision transformer ViT-Base-32~\cite{dosovitskiy2020image}) and the optimal partitioning varies across hardware platforms and other operation configurations (e.g., convolution kernel size), exhaustively measuring latency for every possible partitioning is costly.
Thus, existing works~\cite{jia2022codl,kim2019mulayer,wei2023nn} typically use ML-based latency predictors that rely on the operation parameters (e.g., input/output dimensions).
However, as detailed in \cref{section:gpu_characteristic}, these approaches can hardly capture the complex characteristics of GPU performance, resulting in suboptimal partitioning decisions.
Additionally, as presented in \cref{section:sync_overhead}, synchronization overhead $T_{overhead}$ must be accounted for when evaluating the achieved speedup, since significant overhead can diminish performance gains achieved through co-execution.

%% file: sections/3-predictor.tex
\section{Accurate Latency Prediction}\label{section:latency_prediction}

In this section, we present our approach to improve latency predictors used for partitioning decisions.

\subsection{GPU Kernel Characterization}\label{section:gpu_characteristic}

\begin{figure}[t]
    \centering
    \includegraphics[width=\linewidth]{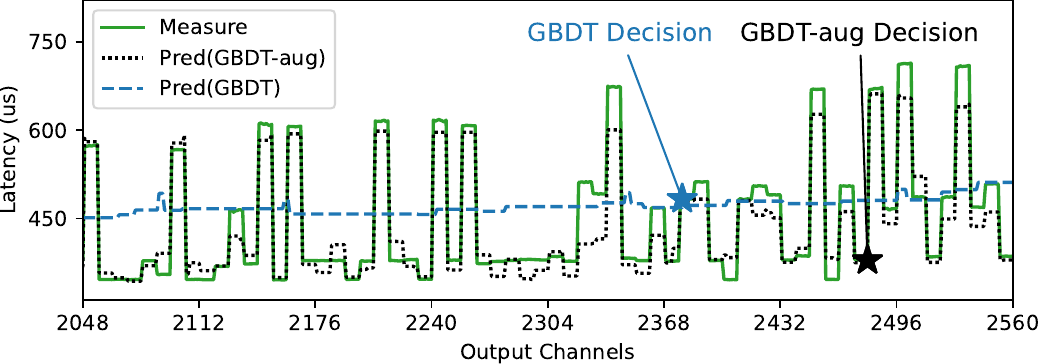}
    \caption{Latency Prediction Improvement using Additional Features  (OnePlus~11)}\label{fig:prediction_improvement}
\end{figure}

\begin{figure}[t]
    \begin{subfigure}[t]{\linewidth}
        \hspace{2.7em}
        \centering
        \includegraphics[width=.745\linewidth]{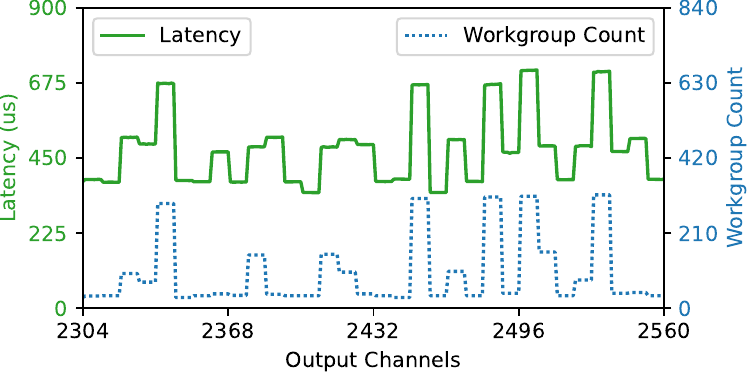}
        \caption{Heuristic Workgroup for Linear Operations with Input Shape (50, 768)}\label{fig:discontinuity_workgroup}
        \vspace{1.5em}
    \end{subfigure}
    \\
    \begin{subfigure}[t]{\linewidth}
        \centering
        \includegraphics[width=.7\linewidth]{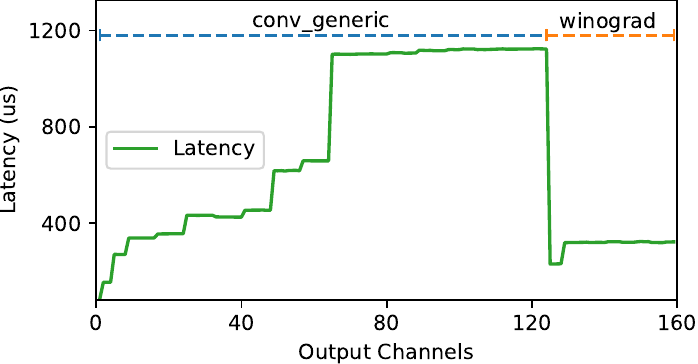}
        \caption{Kernel Switch for 3x3 Convolution Operations with Input Shape (64, 64, 128)}\label{fig:discontinuity_algorithm}
    \end{subfigure}
    \caption{Reasons for Discontinuity in Latency Curve on Mobile GPUs  (OnePlus~11)}\label{fig:motivation_discontinuity}
    \vspace{-1em}
\end{figure}

%
As motivated in \cref{section:background_partition}, accurate latency prediction is crucial for effective partitioning decisions.
To illustrate this, we conduct an experiment on a linear operation $\matr{W}\in\mathbb{R}^{768\times 3072}$ from the vision transformer ViT-Base-32~\cite{dosovitskiy2020image}, where it is used to transform an input feature map $\matr{X}\in\mathbb{R}^{50\times 768}$ into $\matr{Y}\in\mathbb{R}^{50\times 3072}$.
To partition this operation, we measure the GPU latency of linear operations with size $768\times C_{out}$ for $C_{out}\in [2048,2560]$.
As depicted in \cref{fig:prediction_improvement}, GPU latency exhibits significant spikes, e.g., the linear operation with $C_{out}=2500$ is counterintuitively 1.85 times slower than the one with $C_{out}=2520$.
%
Notably, a gradient-boosted decision tree (GBDT, an ML model broadly used in prior work~\cite{LiPG23,LiPG24b,jia2022codl,zhang2021nn}) using matrix sizes as input features only captures the overall increasing trend of latency (blue curve) rather than its sudden changes;
using GBDT latency predictions, 2,378 output channels are assigned to the GPU (measured latency of 483~$\mu$s) and 694 channels are assigned to the CPU (measured latency of 424~$\mu$s), achieving only 1.02x speedup with respect to using only the GPU (as a reference, our approach in  \cref{section:feature_augmentation} achieves 1.29x speedup by assigning 2,480 channels to the GPU, based on more accurate predictions). 
To understand the discontinuity in GPU performance, we analyzed the source code of the ML framework TFLite~\cite{lee2019device} and identified two primary factors accounting for sudden latency changes:
\begin{enumerate}[wide,labelindent=0pt]
\item \emph{Heuristic Workgroup Choices.}
In GPU computing, a workgroup is a collection of threads (work items) concurrently executing the same code and sharing resources such as local memory; the workgroup size determines how GPU threads are grouped for scheduling and is crucial for the efficiency of a GPU kernel.
ML frameworks typically decide the workgroup size heuristically based on the kernel implementation and on the underlying hardware architecture.
\cref{fig:discontinuity_workgroup} presents the latencies and workgroup counts for linear operations with input size $50 \times 768$ and varying number of output channels~$C_{out}$; here, we observe a strong correlation between the number of workgroups and kernel latency.
%

\smallskip\item 
\emph{Kernel Selection.}
In order to enhance performance, TFLite provides multiple GPU kernel implementations of convolution operations for different parameters (kernel size/stride/channels, input/output size).
For example, \cref{fig:discontinuity_algorithm} shows that, for a convolutional layer with $3\times 3$ filter and input size of $64 \times 64 \times 128$, when the number of output channels exceeds 128, the kernel implementation will switch to the Winograd algorithm, which offers higher efficiency for operations with more output channels.
This change of kernel implementations results in substantial latency anomalies due to their distinct performance characteristics.

\end{enumerate}


\subsection{Feature Augmentation}\label{section:feature_augmentation}




To accurately predict the GPU latency, we propose a white-box approach to capture the two aforementioned factors.
Specifically, we analyzed the algorithms of TFLite to determine kernel implementation and workgroup size for a given operation configuration.
Below, we summarize our analysis.

First, TFLite mainly uses three kernel implementations for convolution operations: (1) \texttt{conv\_constant}, which leverages the faster on-chip constant memory when sufficient registers (estimated based on output channels) are available and convolution filters can fit within the constant memory, (2) \texttt{winograd}, which reduces the number of multiplications when the filter size is small and input sizes are large enough to make the transformation overhead negligible compared to the savings in multiplications, and (3) \texttt{conv\_generic}, which is the default implementation for general use cases.
%
Second, once a kernel implementation is selected, ML frameworks heuristically select workgroup configurations by considering hardware-specific factors, such as number of registers, compute unit occupancy, and memory access patterns.
%

To leverage this additional information about GPU kernels, we (1)~construct separate latency predictors for each kernel implementation, and (2)~augment the predictor input features to include kernel dispatch information, such as size and number of workgroups; these dispatch-related features can be calculated based on the hardware specification and on the parameters of the operation being executed.
\cref{fig:prediction_improvement} illustrates that our approach (black curve) accurately captures the spikes of inference latency through feature augmentation.
As a result, our enhanced latency prediction enables more effective workload partitioning for co-execution: using 2,480 output channels on GPU (with measured latency of 379~$\mu$s and prediction improved from 481~$\mu$s to 375~$\mu$s) and 592 output channels on CPU (with measured latency of 354~$\mu$s), the speedup improves from 1.02x to 1.29x (i.e., from 495~$\mu$s to 393~$\mu$s).

%% file: sections/4-overhead.tex
\section{Reduction of CPU-GPU Synchronization Overhead}\label{section:sync_overhead}




%
As discussed in \cref{section:background_partition}, significant synchronization overhead can diminish the performance gains of co-execution.
We identify two main sources of overhead.
\begin{enumerate}[wide,labelindent=0pt]
\item \emph{Data Mapping for Cache Coherence.}
When CPU and GPU collaboratively compute the output of a neural network layer, data transfers between their memory hierarchies are necessary.
In particular, since mobile CPU and GPU have distinct caches, explicit data mapping operations are required to maintain cache coherence.
Previous works \cite{jia2022codl} report up to 1~ms data mapping overhead.

\smallskip
\item \emph{Inter-Processor Notifications.}
Since CPU and GPU computations proceed in parallel, synchronization points are required to manage data dependencies and enforce execution order.
For example, in OpenCL, synchronization between the host (CPU) and device (GPU) can be implemented by making the GPU kernel dependent on a user event; once the CPU finishes its computation, it marks this event as completed.
However, there is a delay before the GPU recognizes the updated event state.
Similarly to earlier work~\cite{jia2022codl}, our measurements indicate that the delay is on average 162~$\mu$s on Motorola Edge Plus 2022 across 2,039 linear operations, which accounts for 39.9\% of the total co-execution latency.
\end{enumerate}

%
%
To reduce synchronization overhead, we use two techniques:
\begin{enumerate}[wide,labelindent=0pt]
\item During inference, we store layer outputs in OpenCL \textit{fine-grained shared virtual memory} (SVM), allowing both the OpenCL host (CPU) and device (GPU) to read and write directly to the same region in main memory. 
For instance, when a previous layer is evaluated collaboratively by CPU and GPU, the output results can be saved to this shared memory and read by subsequent CPU and GPU operations without additional copies between CPU and GPU buffers in main memory.
In addition, unlike \textit{coarse-grained} SVM, \textit{fine-grained} SVM does not require explicit data mapping and unmapping operations since the hardware guarantees cache coherence for memory shared between CPU and GPU.
Hence, this technique not only avoids copying data between CPU and GPU, but also eliminates the cost of data mapping operations.
\smallskip
\item  To reduce the notification delay, we dispatch an active polling OpenCL kernel after each GPU computation.
This kernel updates two synchronization variables (\texttt{cpu\_flag} and \texttt{gpu\_flag}) stored in fine-grained SVM.
The kernel first updates \texttt{gpu\_flag} to indicate GPU completion, then repeatedly checks \texttt{cpu\_flag} to wait for CPU completion.
Meanwhile, the CPU updates \texttt{cpu\_flag} once finishing the computation and keeps polling for \texttt{gpu\_flag} to be updated by the GPU.
Notably, in order to minimize synchronization overhead, our polling-based implementation requires \emph{busy waiting} on both CPU and GPU, which leads to additional power consumption in the case of unbalanced workload partitioning.
However, our accurate latency predictions help balance CPU and GPU computation times and mitigate this issue.
\end{enumerate}
\noindent
Overall, by eliminating data mapping operations and reducing notification delay, we reduce the mean synchronization overhead to 7~$\mu$s on Motorola Edge Plus 2022 across 2,039 linear operations, significantly improving the efficiency of CPU-GPU co-execution.

%% file: sections/5-results.tex
\section{Experimental Results}\label{section:evalution}

In this section, we conduct comprehensive evaluations of our co-execution strategy using 2,039 linear and 2,051 convolution operations on four mobile platforms: Pixel 4, Pixel 5, Motorola Edge Plus 2022 (Moto 2022), and OnePlus 11.

\subsection{Experimental Setup}\label{section:experimental_setup}

Building on TFLite, we developed a C++ benchmarking tool that co-executes OpenCL kernels from the TFLite GPU Delegate~\cite{lee2019device} and CPU kernels from the XNNPACK library~\cite{xnnpack}; CPU-GPU synchronization kernels were implemented in OpenCL and dispatched to the same GPU queue. The compiled binary was uploaded to each smartphone to benchmark the latency of co-execution.

To ensure stable performance measurements, we prepared the smartphones using the same configurations as \cite{LiPG24a}.
Specifically, we enabled the performance mode on the GPU and all CPU cores to encourage near-maximum clock frequencies, reducing performance fluctuations due to dynamic frequency scaling on Android platforms.
The CPU threads were scheduled to high-performance cores by specifying CPU affinity.
In addition, we attached an external cooling fan to the back of the smartphones and allowed for a cool-down time (1 second) after profiling each operation configuration, in order to mitigate thermal throttling effects, which can significantly hinder sustained performance.

\subsection{Training Dataset Generation and Latency Prediction Accuracy}\label{section:evalution_setup}

We construct a training dataset for ML predictors by sampling a broad range of operation parameters.
For linear layers, operation dimensions (input length $L$, input channels $C_{in}$, and output channels $C_{out}$) are selected using a structured random sampling approach:  first, we randomly pick an interval from $\{ [2^k, 2^{k+1}] \mid \ 2 \le k \le 9\}$, and then we sample the dimensions uniformly from the selected interval.
For convolutional layers, we sample input height~$H_{in}$, input width~$W_{in}$, input channels~$C_{in}$, and output channels $C_{out}$ using the same approach, and we sample the kernel (filter) shape $K$ from $\{1,3,5,7\}$ and its stride $S$ from $\{1,2\}$.
In total, we collect latency measurements for 12,500 distinct configurations for each type of layer (linear or convolutional), using $20\%$ for testing.

\begin{figure}[t]
    \centering\includegraphics[width=\linewidth]{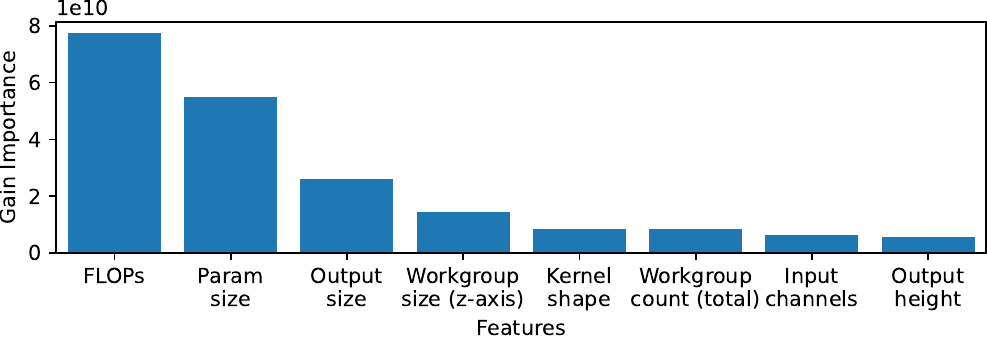}\vspace{-.5em}
    \caption{Gain Improvement from GBDT Input Features (Convolution, Moto 2022)}\label{fig:GBDT_gain_importance}
\end{figure}

To train latency predictors, we use gradient-boosted decision trees (GBDTs) from LightGBM~\cite{ke2017lightgbm} 
and Optuna~\cite{akiba2019optuna} to tune hyperparameters, which include learning rate (0.01 to 0.2), number of estimators (100 to 1000), depth (5 to 20), number of leaves (16 to 512), L1/L2 regularization terms ($10^{-8}$ to 1), and subsample ratios (0.5 to 1).
%
%
As discussed in \cref{section:feature_augmentation}, we train separate predictors for each kernel implementation, based on features including operation configurations and workgroup information.
\cref{fig:GBDT_gain_importance} presents the gain importance (i.e., the total loss improvement for all splits of a feature) for the top eight features of convolutional layers.
As shown, workgroup size and total workgroup count are important factors affecting latency, which motivates their inclusion as input features. 
The GBDT predictors typically take 3-4~ms to determine the optimal partitioning for each operation; these partitioning decisions can be made offline before deployment to a device, as part of the compilation process.

%

\begin{table}[t]
\centering
\setlength{\tabcolsep}{0.5em} 
\renewcommand\arraystretch{1.1}
\begin{tabular}{c c c c c c}\toprule
\multirow{2}{*}{\shortstack{\\\textbf{Device}}} & \multirow{2}{*}{\shortstack{\\\textbf{Operations}}} & \multicolumn{4}{c}{\textbf{MAPEs}} \\
\cmidrule(lr){3-6}
& & GPU & 1 CPU & 2 CPUs & 3 CPUs \\
\toprule

\multirow{2}{*}{Pixel 4}
& Linear & 4.4\% & 11.5\% & 7.1\% & 5.8\% \tabularnewline
& Convolutional & 8.5\% & 11.4\% & 8.8\% & 7.2\% \tabularnewline
\midrule

\multirow{2}{*}{Pixel 5}
& Linear & 3.7\% & 6.2\% & 7.8\% & 7.2\% \tabularnewline
& Convolutional & 7.7\% & 6.9\% & 8.1\% & 7.1\% \tabularnewline
\midrule

\multirow{2}{*}{Moto 2022}
& Linear & 4.0\% & 2.5\% & 2.6\% & 2.4\% \tabularnewline
& Convolutional & 9.0\% & 4.0\% & 3.6\% & 3.5\% \tabularnewline
\midrule

\multirow{2}{*}{OnePlus 11}
& Linear & 3.7\% & 3.1\% & 2.9\% & 3.1\% \tabularnewline
& Convolutional & 7.4\% & 4.8\% & 4.2\% & 4.4\% \tabularnewline

\bottomrule
\end{tabular}
\vspace{2mm}
\caption{MAPEs of GBDT Predictors}
\label{table:gbdt_mape}
\end{table}

\cref{table:gbdt_mape} presents the prediction Mean Average Percentage Error (MAPE); errors are generally higher for convolutions due to the greater number of parameters (e.g., filter shape, stride) and multiple kernel implementations.

\begin{table}[t]
\centering
\setlength{\tabcolsep}{0.3em} 
\renewcommand\arraystretch{1.1}
\begin{tabular}{c c c c c c c c}\toprule
\multirow{2}{*}{\shortstack{\\\textbf{Device}}} & \multirow{2}{*}{\shortstack{\\\textbf{Method}}} & \multicolumn{3}{c}{\textbf{Speedup of Linear}} & \multicolumn{3}{c}{\textbf{Speedup of Convolutional}} \\
\cmidrule(lr){3-5} \cmidrule(lr){6-8}
& & 1 thread & 2 threads & 3 threads & 1 thread & 2 threads & 3 threads \\
\toprule

\multirow{2}{*}{Pixel 4}
 & GBDT & 1.21x & 1.52x & 1.84x & 1.22x & 1.46x & 1.69x \tabularnewline
 & Search & 1.29x & 1.59x & 1.92x & 1.31x & 1.56x & 1.79x \tabularnewline
\midrule

\multirow{2}{*}{Pixel 5}
 & GBDT & 1.51x & 1.78x & 1.89x & 1.45x & 1.69x & 1.75x \tabularnewline
 & Search & 1.63x & 1.92x & 2.01x & 1.49x & 1.80x & 1.87x \tabularnewline
\midrule

\multirow{2}{*}{Moto 2022}
 & GBDT & 1.20x & 1.32x & 1.44x & 1.16x & 1.27x & 1.39x \tabularnewline
 & Search & 1.23x & 1.36x & 1.49x & 1.22x & 1.34x & 1.46x \tabularnewline
\midrule

\multirow{2}{*}{OnePlus 11}
 & GBDT & 1.06x & 1.17x & 1.26x & 1.07x & 1.22x & 1.35x \tabularnewline
 & Search & 1.13x & 1.25x & 1.35x & 1.12x & 1.27x & 1.40x \tabularnewline

\bottomrule
\end{tabular}
\vspace{2mm}
\caption{Average Speedups Obtained through CPU-GPU Co-Execution}
\label{table:speedup_summary}
\end{table}

\subsection{Co-Execution Speedup of Individual Layers}\label{section:evalution_speedup}

%
\cref{table:speedup_summary} reports the speedups of our co-execution strategy, together with the best speedups found by a grid search over $[0,C_{out}]$ with step size equal to 8.
Given the long times required for repeated measurements, grid search is evaluated only on a random subset including 10\% of the test cases. Note that grid search is only a baseline that is not applicable to real-world systems due to the long measurement times required for each new set of parameters of the linear and convolution operations.
Test cases are selected as follows.

\begin{itemize}[wide,labelindent=0pt]
\item \emph{Linear Layers:} Dimensions are selected from $\{i \cdot 2^j\ | \ 4 \le i \le 6, 2 \le j \le 9\}$.
Then, we keep operations with FLOPs in  $[4 \cdot 10^6, 10^9]$, resulting in a total of 2,039 linear operations.
Our predictor leads to up to 1.89x average speedup (on Pixel~5), close to the best average speedup of 2.01x obtained through grid search.

\smallskip
\item \emph{Convolutional Layers:} Mobile vision models adopt hierarchical stages~\cite{sandler2018mobilenetv2}, with earlier stages capturing the local details (e.g., texture) and later stages capturing global features (e.g., shape).
Accordingly, we define a hierarchy of 4 stages:
convolutions of the first stage have input resolution $H_{in}, W_{in}$ $\in \{64, 56, 48, 40\}$, kernel (filter) shape $K\in\{1, 3, 5, 7\}$, stride $S\in\{1, 2\}$, and input/output channels $C_{in}, C_{out} \in \{256/i, 320/i, 384/i, 448/i, 512/i\}$ where $i=1,1,4,8$ for $K=1, 3, 5, 7$, respectively (to have similar computation loads).
In stages 2, 3, 4, we halve the resolution and double the number of channels.
We keep the generated layers with FLOPs within range $[4\cdot 10^6, 10^9]$, resulting in a total of 2,051 convolutional layers.
As shown in \cref{table:speedup_summary}, our predictor achieves up to 1.75x speedup on Pixel~5, close to the best measured speedup of 1.87x.
Notably, speedups are higher on devices with smaller CPU/GPU performance gap (e.g., Pixel 4 and Pixel 5), as a larger fraction of the task can be offloaded to the CPU.
\end{itemize}

\subsection{Co-Execution Speedup of End-to-End Models}

\begin{table}[t]
\centering
\setlength{\tabcolsep}{0.3em} 
\renewcommand\arraystretch{1.1}
\begin{tabular}{c c c c c c c}\toprule
\multirow{2}{*}{\shortstack{\\\textbf{Device}}} & \multirow{2}{*}{\shortstack{\\\textbf{Neural}\\\textbf{Network}}} & \multirow{2}{*}{\shortstack{\\ \textbf{Baseline} \\ \textbf{(ms)}}} & \multicolumn{2}{c}{\textbf{Individual Ops}} & \multicolumn{2}{c}{\textbf{End-to-End}} \\
\cmidrule(lr){4-5} \cmidrule(lr){6-7}
& & & Latency~(ms) & Speedup & Latency~(ms) & Speedup \\
\toprule

\multirow{4}{*}{Pixel 4}
 & VGG16 & \phantom{0}83.3 & \phantom{0}70.8 & 1.18x & \phantom{0}73.0 & 1.14x \tabularnewline
 & ResNet-18 & \phantom{0}17.5 & \phantom{0}11.0 & 1.59x & \phantom{0}11.4 & 1.54x \tabularnewline
 & ResNet-34 & \phantom{0}37.5 & \phantom{0}22.2 & 1.69x & \phantom{0}22.5 & 1.67x \tabularnewline
 & Inception-v3 & \phantom{0}99.5 & \phantom{0}59.1 & 1.68x & \phantom{0}61.5 & 1.62x \tabularnewline
\midrule

\multirow{4}{*}{Pixel 5}
 & VGG16 & 194.8 & 119.8 & 1.63x & 125.1 & 1.56x \tabularnewline
 & ResNet-18 & \phantom{0}33.2 & \phantom{0}18.2 & 1.82x & \phantom{0}18.6 & 1.78x \tabularnewline
 & ResNet-34 & \phantom{0}65.2 & \phantom{0}36.9 & 1.77x & \phantom{0}37.1 & 1.76x \tabularnewline
 & Inception-v3 & 184.3 & \phantom{0}99.9 & 1.85x & 102.9 & 1.79x \tabularnewline
\midrule

\multirow{4}{*}{Moto 2022}
 & VGG16 & \phantom{0}32.0 & \phantom{0}28.7 & 1.11x & \phantom{0}29.7 & 1.08x \tabularnewline
 & ResNet-18 & \phantom{00}7.5 & \phantom{00}6.3 & 1.18x & \phantom{00}6.7 & 1.11x \tabularnewline
 & ResNet-34 & \phantom{0}14.7 & \phantom{0}12.2 & 1.20x & \phantom{0}12.9 & 1.14x \tabularnewline
 & Inception-v3 & \phantom{0}52.0 & \phantom{0}38.8 & 1.34x & \phantom{0}41.1 & 1.27x \tabularnewline
\midrule

\multirow{4}{*}{OnePlus 11}
 & VGG16 & \phantom{0}27.4 & \phantom{0}25.3 & 1.09x & \phantom{0}26.2 & 1.05x \tabularnewline
 & ResNet-18 & \phantom{00}8.5 & \phantom{00}6.6 & 1.29x & \phantom{00}6.8 & 1.25x \tabularnewline
 & ResNet-34 & \phantom{0}17.6 & \phantom{0}13.5 & 1.31x & \phantom{0}13.8 & 1.27x \tabularnewline
 & Inception-v3 & \phantom{0}44.2 & \phantom{0}36.2 & 1.22x & \phantom{0}37.8 & 1.17x \tabularnewline

\bottomrule
\end{tabular}
\vspace{2mm}
\caption{End-to-End Speedups from GPU and 3 CPU Threads Co-Execution}
\label{table:e2e_experiments}
\end{table}

In this section, we present experiments demonstrating the speedups in end-to-end latency.
We focus on four common neural networks: VGG16~\cite{vgg}, ResNet-18~\cite{resnet}, ResNet-34, and Inception-v3~\cite{inception-v3}.
In \cref{table:e2e_experiments}, we report the baseline latency for running each model on GPU.
As comparison, we evaluate the performance of individual convolution and linear operations through co-execution on GPU and 3 CPU threads;
we also conduct end-to-end experiments that incorporate the offline partitioning decisions for each operation and correspondingly schedule CPU and GPU kernels for the entire model; notably, pooling operations are always scheduled on the GPU, since their latency is negligible and this can avoid the synchronization overhead.
The end-to-end improvement is slightly lower than that of individual operations, potentially due to memory access overhead between layers.
The results show that our proposed method can achieve up to 1.67x, 1.79x, 1.27x, and 1.27x average speedups on Pixel 4, Pixel 5, Moto 2022, and OnePlus 11, respectively; these results validate the effectiveness of approach to real-world neural networks.

We observe that related work~\cite{jia2022codl} also evaluated co-execution of VGG16, reducing its inference latency on Pixel~4 from a baseline of 200~ms (using only the GPU) to around 150~ms (using CPU and GPU).
In contrast, our evaluation of VGG16 on Pixel~4 started from a baseline of 83.3~ms, which was reduced to 73.0~ms through co-execution.
The difference in baseline inference latency is due to the different performance of the MACE ML framework~\cite{mace} (used in \cite{jia2022codl}) and TFLite (used in our work).
In particular, TFLite (1) already leverages image storage types to take advantage of L1 texture cache (which was used in \cite{jia2022codl} to improve the MACE baseline) and (2) implements efficient Winograd kernels to accelerate convolutional layers in VGG16~(\cref{section:feature_augmentation}).


\subsection{Ablation Study}\label{section:evaluation_ablation_study}

\begin{table}[t]
\centering
\renewcommand\arraystretch{1.1}
\begin{tabular}{c c c c c c c}\toprule
\multirow{2}{*}{\shortstack{\\\textbf{Method}}} & \multicolumn{3}{c}{\textbf{Speedup of Linear}} & \multicolumn{3}{c}{\textbf{Speedup of Convolutional}} \\
\cmidrule(lr){2-4} \cmidrule(lr){5-7}
& 1 thread & 2 threads & 3 threads & 1 thread & 2 threads & 3 threads \\
\toprule

Ours & \phantom{\hspace{3.5mm}}1.20x\phantom{\hspace{3.5mm}} & \phantom{\hspace{3.5mm}}1.32x\phantom{\hspace{3.5mm}} & \phantom{\hspace{3.5mm}}1.44x\phantom{\hspace{3.5mm}} & \phantom{\hspace{3.5mm}}1.16x\phantom{\hspace{3.5mm}} & \phantom{\hspace{3.5mm}}1.27x\phantom{\hspace{3.5mm}} & \phantom{\hspace{3.5mm}}1.39x\phantom{\hspace{3.5mm}} \tabularnewline
w/o Augmentation & 1.12x & 1.24x & 1.37x & 1.08x & 1.19x & 1.31x \tabularnewline
Original Overhead & 0.76x & 0.81x & 0.88x & 0.98x & 1.07x & 1.17x \tabularnewline

\bottomrule
\end{tabular}
\vspace{2mm}
\caption{Ablation Study: Co-execution Speedup (Moto 2022)}
\label{table:ablation_study}
\end{table}

We conduct an ablation study on Moto 2022 to evaluate the individual impact of our proposed techniques.
%
%
First, we observe that our augmentation technique reduces latency prediction MAPE of linear layers from 9.3\% to 4.4\% and of convolutional layers from 14.1\% to 9.3\%.
These improvements lead to better partitioning strategies, e.g., for convolutional layers using the GPU and 1 CPU thread, the latency reduction improves from 1.08x to 1.16x (\cref{table:ablation_study}).

Next, to evaluate the contribution of our overhead reduction technique, we compare our active polling implementation with a baseline where the CPU \emph{passively} waits for  GPU kernel completions using the OpenCL \texttt{clWaitForEvents} API.
The baseline incurs average overhead of 162~$\mu$s across 2,039 linear layers and 141~$\mu$s across 2,051 convolutional layers, accounting for 39.9\% and 15.8\% of the total co-execution latency, respectively, in the case of GPU co-execution with 1 CPU thread. Such high overhead negates the benefits of co-execution; in contrast, our active polling approach makes the synchronization overhead negligible (average of 7.0~$\mu$s for linear layers and 5.4~$\mu$s for convolutional layers).

%% file: sections/6-conclusion.tex
\section{Conclusions}\label{section:conclusion}

We explored inference latency optimization for deep neural networks on mobile platforms by partitioning individual linear and convolutional layers across CPU and GPU.
To address challenges in accurately predicting complex GPU latency behaviors and reducing CPU-GPU synchronization overhead, we developed enhanced latency predictors incorporating kernel implementation and dispatching information, and we proposed a lightweight synchronization method using OpenCL shared virtual memory.
Comprehensive experimental evaluations showed that our approach achieves significant speedups that are close to the achievable best.
In future work, we plan to investigate parallel execution on CPU, GPU, and NPU, and the effects of model quantization.

\section*{Acknowledgments}
This work was supported in part by the NSF CNS-1816887, CCF-1763747, and IIS-1833137 awards.
The authors would like to thank the anonymous EPEW reviewers for their insightful comments that helped improve this paper.